\pdfoutput=1

\documentclass[11pt]{article}

\usepackage[final]{acl}

\usepackage{times}
\usepackage{latexsym}
\usepackage{enumitem}

\usepackage[T1]{fontenc}

\usepackage[utf8]{inputenc}

\usepackage{microtype}
\usepackage{array}
\usepackage{graphicx}
\usepackage{subcaption}
\usepackage{booktabs}
\usepackage{amsmath}
\usepackage{multirow,makecell}
\usepackage{lipsum}
\usepackage{color,soul}
\usepackage{pifont}
\definecolor{mygreen}{HTML}{006400}
\definecolor{mypurple}{HTML}{9D3972}

\newcommand{\name}[0]{RELiC}
\newcommand{\claim}[1]{\textcolor{mygreen}{#1}}
\newcommand{\litquote}[1]{\textcolor{mypurple}{#1}}
\newcommand\customfont[1]{{\usefont{T1}{greek}{m}{n} #1 }}
\newcommand{\model}[0]{dense-\name}
\definecolor{mygray}{gray}{0.4}
\newcommand{\cmark}{\color{mygray}\ding{51}}%
\newcommand{\xmark}{\color{mygray}\ding{55}}%
\newcommand{\book}[0]{\includegraphics[width=.025\textwidth]{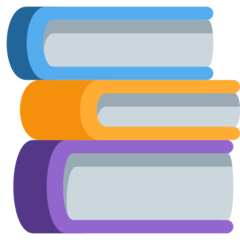}}
\newcommand{\scroll}[0]{\includegraphics[width=.025\textwidth]{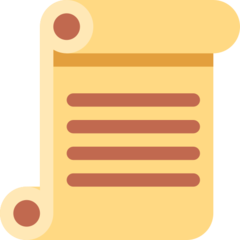}}

\title{{\LARGE \customfont{R\,E\,L\,\,I\,C}}:  Retrieving Evidence for Literary Claims}

\author{\bf Katherine Thai\textsuperscript{\book} \quad Yapei Chang\textsuperscript{\scroll} \quad Kalpesh Krishna\textsuperscript{\book} \quad Mohit Iyyer\textsuperscript{\book} \\\\ \textsuperscript{\book}University of Massachusetts Amherst, \textsuperscript{\scroll}Smith College \\
\texttt{\{kbthai,kalpesh,miyyer\}@cs.umass.edu} \\
\texttt{echang33@smith.edu}\vspace{0.2cm}
\\ \textbf{Project Page}: \texttt{\href{https://relic.cs.umass.edu}{https://relic.cs.umass.edu}}}

\begin{document}
\maketitle
\begin{abstract}
Humanities scholars commonly provide evidence for \claim{claims} that they make about a work of literature (e.g., a novel) in the form of \litquote{quotations} from the work. 
 We collect a large-scale dataset (\name) of 78K literary quotations and surrounding critical analysis and use it to formulate the novel task of \textit{literary evidence retrieval}, in which models are given an \claim{excerpt of literary analysis} surrounding a \litquote{masked quotation} and asked to retrieve the quoted passage from the set of all passages in the work. Solving this retrieval task requires a deep understanding of complex literary and linguistic phenomena, which proves challenging to methods that  overwhelmingly rely on lexical and semantic similarity matching. We implement a RoBERTa-based dense passage retriever for this task that outperforms existing pretrained information retrieval baselines; however, experiments and analysis by human domain experts indicate that there is substantial room for improvement over our dense retriever.
\end{abstract}
\section{Introduction}
\label{sec:introduction}

When analyzing a literary work (e.g., a novel or short story), scholars make \claim{claims} about the text and provide supporting evidence in the form of \litquote{quotations} from the work~\citep{thompson2002death,finnegan2011we,graff2014they}. For example,~\citet{monaghan_jane_1980} claims that Elizabeth, the main character in Jane Austen's \textit{Pride and Prejudice}, doesn't just refuse an offer to join the standoffish bachelor Darcy and the wealthy Bingleys on their morning walk, \claim{``but does so in such a way as to group Darcy with the snobbish Bingley sisters,''} and then directly quotes Elizabeth's tongue-in-cheek rejection: \litquote{``No, no; stay where you are. You are charmingly grouped, and appear to uncommon advantage. The picturesque would be spoilt by admitting a fourth.''}

Literary scholars construct arguments like these by making complex connective inferences between their interpretations, framed as \claim{claims}, and \litquote{quotations} (e.g., recognizing that Elizabeth says \litquote{``charmingly grouped''} and \litquote{``picturesque''} ironically in order to \claim{group Darcy with the snobbish Bingley sisters}). This process requires a deep understanding of both literary phenomena, such as irony and metaphor, and linguistic phenomena (coreference, paraphrasing, and stylistics).  
In this paper, we computationally study the relationship between literary claims and quotations by collecting a large-scale dataset for \textbf{R}etrieving \textbf{E}vidence for \textbf{Li}terary \textbf{C}laims (\name), which contains 78K
scholarly excerpts of literary analysis that each directly quote a passage from one of 79 widely-read English texts. 

\begin{figure*}[ht]
\centering
\includegraphics[width=\linewidth]{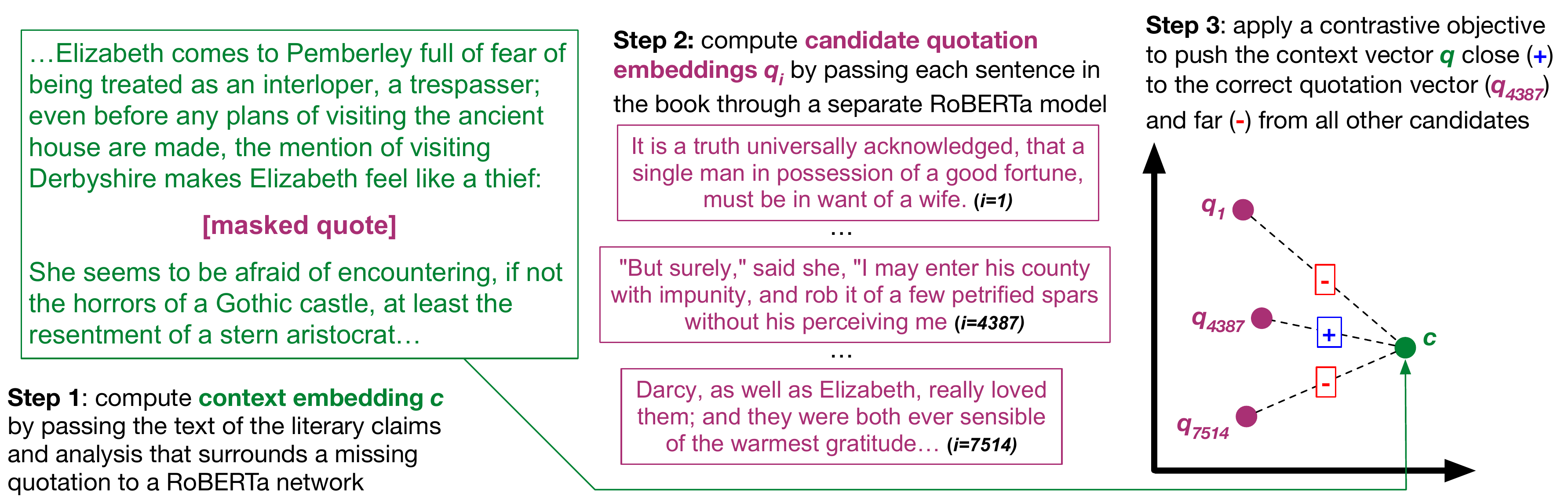}
\caption{An example of our \textit{literary evidence retrieval} task and the model we built to solve it. The model must retrieve a missing \litquote{quotation from \emph{Pride and Prejudice}} given the \claim{literary claims and analysis} that surround the quotation. The retrieval candidate set for this example consists of all 7,514 sentences from \emph{Pride and Prejudice}. Our \model\ model is trained with a contrastive loss to push a learned representation of the surrounding context close to a representation of the ground-truth missing quotation (here, the 4,387$^\text{th}$ sentence from the novel).}
\label{fig:model_example}
\end{figure*}

The complexity of the claims and quotations in \name\ makes it a challenging testbed for modern neural retrievers: given just the \claim{text of the claim and analysis} that surrounds a masked \litquote{quotation}, can a model retrieve the quoted passage from the set of all possible passages in the literary work? This \emph{literary evidence retrieval} task (see Figure~\ref{fig:model_example}) differs considerably from retrieval problems commonly studied in NLP, such as those used for fact checking \cite{thorne-etal-2018-fever}, open-domain QA \cite{chen-etal-2017-reading,chen-yih-2020-open}, and text generation \cite{lfqa21}, in the relative lack of lexical or even semantic similarity between claims and queries. Instead of latching onto surface-level cues, our task requires models to understand complex devices in literary writing and apply general theories of interpretation. \name\ is also challenging because of the large number of retrieval candidates: for \textit{War and Peace}, the longest literary work in the dataset, models must choose from one of $\sim$~32K candidate passages.


How well do state-of-the-art retrievers perform on \name? Inspired by recent research on dense passage retrieval~\citep{guu2020realm,karpukhin-etal-2020-dense}, we build a neural model (\model) by embedding both scholarly claims and candidate literary quotations with pretrained RoBERTa networks~\citep{liu2019roberta}, which are then fine-tuned using a contrastive objective that encourages the representation for the ground-truth quotation to lie nearby to that of the claim. 
Both sparse retrieval methods such as BM25 and pretrained dense retrievers such as DPR and REALM perform poorly on \name, which underscores the difference between our dataset and existing information retrieval benchmarks~\citep{thakur2021beir} on which these baselines are much more competitive. Our \model\ model fares better than these baselines but still lags far behind human performance, and an analysis of its errors suggests that it struggles to understand complex literary phenomena.

Finally, we qualitatively explore whether our \model\ model can be used to support evidence-gathering efforts by researchers in the humanities. Inspired by prompt-based querying~\cite{jiang_how_2020}, we issue our own out-of-distribution queries to the model by formulating simple descriptions of events or devices of interest (e.g., \emph{symbols of Gatsby's lavish lifestyle}) and discover that it often returns relevant quotations. To facilitate future research in this direction, we publicly release our dataset and models.\footnote{\url{https://relic.cs.umass.edu}}

\section{Collecting a Dataset for Literary Evidence Retrieval}
\label{sec:dataset}
We collect a dataset for the task of  \textbf{R}etrieving \textbf{E}vidence for \textbf{Li}terary \textbf{C}laims, or \name, the first large-scale retrieval dataset that focuses on the challenging literary domain. Each example in \name\ consists of two parts: (1) the \claim{context surrounding the quoted material}, which consists of literary claims and analysis, and (2) a \litquote{quotation} from a widely-read English work of literature. This section describes our data collection and preprocessing, as well as a fine-grained analysis of 200 examples from \name\ to shed light on the types of quotations it contains. See Table \ref{tab:corpus_stats} for corpus statistics.

\subsection{Collecting and Preprocessing \name}
\paragraph{Selecting works of literature:}
We collect 79 primary source works written or translated into English\footnote{Of the 79 primary sources in \name, 72 were originally written in English, 3 were written in French, and 4 were written Russian. \name\ contains the corresponding English translations of these 7 primary source works. The complete list of primary source works is available in Appendix Tables \ref{appendix:training}, \ref{appendix:val_test}.}  from Project Gutenberg and Project Gutenberg Australia.\footnote{\url{https://www.gutenberg.org/}} These public domain sources were selected because of their popularity and status as members of the Western literary canon, which also yield more scholarship~\cite{porter_lit_lab}.
All primary sources were published in America or Europe between 1811 and 1949. 77 of the 79 are fictional novels or novellas, one is a collection of short stories (\textit{The Garden Party and Other Stories} by Katherine Mansfield), and one is a collection of essays (\textit{The Souls of Black Folk} by W. E. B. Du Bois). 

\paragraph{Collecting quotations from literary analysis:} We queried all documents in the HathiTrust Digital Library,\footnote{\url{https://www.hathitrust.org/}} a collaborative repository of volumes from academic and research libraries, for exact matches of all sentences of ten or more tokens from each of the 79 works. The overwhelming majority of HathiTrust documents are scholarly in nature, so most of these matches yielded critical analysis of the 79 primary source works. We received permission from the HathiTrust to publicly release short windows of text surrounding each matching quotation. 

\paragraph{Filtering and preprocessing:} The  scholarly articles we collected from our HathiTrust queries were filtered to exclude duplicates and non-English sources. We then preprocessed the resulting text to remove pervasive artifacts such as in-line citations, headers, footers, page numbers, and word breaks using a pattern-matching approach (details in Appendix~\ref{appendix:data}). Finally, we applied sentence tokenization using spaCy's dependency parser-based sentence segmenter\footnote{\url{https://spacy.io/}, the default segmenter in spaCy is modified to use ellipses, colons, and semicolons as custom sentence boundaries, based on the observation that literary scholars often only quote part of what would typically be defined as a sentence.} to standardize the size of the windows in our dataset. 
Each window in \name\ contains the \litquote{identified quotation} and \claim{four sentences of claims and analysis}\footnote{The HathiTrust permitted us to release windows consisting of up to eight sentences of scholarly analysis. While more context is of course desirable, we note that (1) conventional model sizes are limited in input sequence length, and (2) context further away from the quoted material has diminishing value, as it is likely to be less relevant to the quoted span.} \claim{on each side of the quotation} (see Table~\ref{tab:dataset_examples} for examples). To avoid asking models to retrieve a quote they have already seen during training, we create training, validation, and test splits such that primary sources in each fold are mutually exclusive. Statistics of our dataset sources are provided in Appendix \ref{sec:more-data-stats}.


\begin{table}
\small

\begin{center}
\begin{tabular}{ l r }  
 \toprule
 \# training examples & 62,956\\
 \# validation examples &  7,833 \\
 \# test examples & 7,785\\
 \# total examples & 78,574\\
 \midrule
 average \claim{context} length (words) & 157.7 \\
 average \litquote{quotation} length (words) & 40.5 \\
 \midrule
 \# primary sources & 79\\
 \# unique sec. sources & 8,836\\
 \bottomrule
 
\end{tabular}

\end{center}
\caption{\label{tab:corpus_stats}\name\ statistics. Primary sources are from Project Gutenberg and Project Gutenberg Australia. Secondary sources are from the HathiTrust.}
\end{table}


 


\subsection{Comparison to other retrieval datasets}
\label{sec:compare-other-datasets}

Table \ref{tab:corpus_stats} contains detailed statistics of \name. To the best of our knowledge, \name\ is the first retrieval dataset in the literary domain, and the only one that requires understanding complex phenomena like irony and metaphor. We provide a detailed comparison of \name\ to other retrieval datasets in the recently-proposed BEIR retrieval benchmark~\citep{thakur2021beir} in Appendix Table~\ref{tab:compare-beir}. \name\ has a much longer query length (157.7 tokens on average) than all BEIR datasets except ArguAna~\citep{wachsmuth:2018a}. Furthermore, our results in Section \ref{sec:results} show that while these longer queries confuse pretrained retriever models (which heavily rely on token overlap), a model trained on \name\ is able to leverage the longer queries for better retrieval.

\begin{table*}[ht]
\footnotesize
  \centering
  \begin{tabular}{m{2cm} m{13cm}}
    \toprule
    \textbf{Quote type} & \claim{Preceding context}, \litquote{\textbf{primary source quotation}}, \claim{subsequent context} \\
    \midrule
    Claim-supporting evidence (\emph{153})& \claim{If this whale inspires the most lyrical passages in the novel, it also brings into focus such fundamental questions as the knowability of space:} \litquote{\textbf{And as for this whale spout, you might almost stand in it, and yet be undecided as to what it is precisely.}} \claim{But Ishmael stands before the paradoxes of reality with historical and scientific intellect, wisdom, and comic elasticity that accommodates--however tenuously--the uncertainties of this world \cite{hartstein_myth_1985}.}\\
    \midrule
    Paraphrase-supporting evidence (\emph{25}) &  \claim{But then, suddenly, Jacob's thought switches back to the lantern under the tree, with the old toad and the beetles and the moths crossing from side to side in the light, senselessly.} \litquote{\textbf{Now there was a scraping and murmuring.  He caught Timmy Durrant's eye;  looked very sternly at him;  and then, very solemnly, winked.}} \claim{From a boat on the Cam there is another sort of beauty to be seen. There are buttercups gilding the meadows, and cows munching, and the legs of children deep in the grass. Jacob looks at all these things and becomes absorbed \cite{blackstone_virginia_1972}.}\vspace{0.1cm} \\
    \toprule
        Claim-supporting evidence & \claim{The relationship between Alexandra and the earth is an intensely personal one:} \litquote{\textbf{For the first time, perhaps, since that land emerged from the waters of geologic ages, a human face was set toward it with love and yearning...}} \claim{The religious connotations of the more lyrical descriptions of the land prepare us for the emergence of Alexandra as its goddess~\citep{helmick_myth_1968}.}\\
    \midrule
    Paraphrase-supporting evidence & \claim{O Pioneers! is the story of a Swedish immigrant, Alexandra Bergson, who some to Nebraska with her parents when she is young. Her father dies, and she has to take over the farm and look after her younger brothers. Her courage, vision, and energy bring life and civilization to the wilderness. As Alexandra faces the future after her father's death, Willa Cather writes:} \litquote{\textbf{For the first time, perhaps, since that land emerged from the waters of geologic ages, a human face was set toward it with love and yearning.}} \claim{The history of every country begins in the heart of a man or a woman. Alexandra succeeds in taming the wild land, and after a heaping measure of material success and personal tragedy, she faces the future calmly.~\citep{woodress_willa_1975}.}\\
    \bottomrule
  \end{tabular}
  \caption{Examples of the two major types of evidence identified in our manual analysis of \name. \emph{Claim-supporting} evidence uses quotations to support more general literary claims, while \emph{paraphrase-supporting evidence} uses quotations to corroborate summaries of the plot. The bottom two rows show the same quotation (from Willa Cather's \textit{O Pioneers!}) being used as evidence in different ways, highlighting the dataset's complexity.}
  \label{tab:dataset_examples}
\end{table*}

\subsection{Analyzing different types of quotation} What are the different ways in which literary scholars use direct quotation in \name? We perform a manual analysis of 200 held-out examples to gain a better understanding of quotation usage, categorizing each quotation into the following three types:


\paragraph{Claim-supporting evidence:} In 151 of the 200 annotated examples, literary scholars used direct quotation to provide evidence for a more general claim about the primary source work. In the first row of Table \ref{tab:dataset_examples}, \citet{hartstein_myth_1985} claims that \claim{``this whale... brings into focus such fundamental questions as the knowability of space:''}
and then quotes the following metaphorical description from \emph{Moby Dick} as evidence: \litquote{``And as for this whale spout, you might almost stand in it, and yet be undecided as to what it is precisely.''}
When quoted material is used as \textbf{claim-supporting evidence}, the context before and after usually refers directly to the quoted material;\footnote{In 19 of the 151 \textbf{claim-supporting evidence} examples,  scholars  introduce quoted material by explicitly referring to a specific ``sentence,'' ``passage,'' ``scene,'' or similar delineation.} for example, the \claim{paradoxes of reality} and \claim{uncertainties of this world} are exemplified by the vague nature of the whale spout.

\paragraph{Paraphrase-supporting evidence:} In 31 of the examples, we observe that scholars used the primary source work to support their own paraphrasing of the plot in order to contextualize later analysis. In the second row of Table \ref{tab:dataset_examples}, \citet{blackstone_virginia_1972} uses the quoted material to enhance a summary of a specific scene in which Jacob's mind is wandering during a chapel service. Jacob's daydreaming is later used in an analysis of Cambridge as a location in Virginia Woolf's works, but no literary argument is made in the immediate context. When quoted material is being employed as \textbf{paraphrase-supporting evidence}, the surrounding context does not refer directly to the quotation.

\paragraph{Miscellaneous:} 18 of the 200 samples were not literary analysis, though some were still related to literature (for example, analysis of the the film adaptation of \textit{The Age of Innocence}). Others were excerpts from the primary sources that suffered from severe OCR artifacts and were not detected or extracted by the methods in Appendix~\ref{appendix:standardization}.

\section{Literary Evidence Retrieval}
\label{sec:experiments}

\begin{table*}[ht]
\small
\begin{center}
\begin{tabular}{ lrrrrrrrrr } 
 \toprule
 \bf Model & \bf \claim{L/R} & \multicolumn{6}{c}{\bf Recall@\emph{k} ($\uparrow$)} & \bf Avg rank ($\downarrow$) & \bf \multirowcell{2}{Proxy task \\ acc ($\uparrow$)} \\
\cmidrule{3-8}\vspace{-0.3cm}\\
  & & 1 & 3 & 5 & 10 & 50 & 100 &  &  \\
 \midrule
 \multicolumn{10}{c}{\textit{(non-parametric / pretrained zero-shot)}}\vspace{0.1cm}\\
 random & & 0.0 & 0.1 & 0.1 & 0.2 & 1.2 & 2.5 & 2445.1 & 33.3 \\
 BM25 & 1/1 & 1.2 & 3.2 & 4.2 & 5.9 & 12.5 & 17.0 & 1561.2 & --\footnotemark[9] \\
 BM25 & 4/4 & 1.3 & 2.9 & 4.1 & 6.7 & 14.5 & 19.7 & 1386.8 & -- \\
 SIM~\citep{wieting-etal-2019-beyond} & 1/1 & 1.3 & 2.8 & 3.8 & 5.6 & 13.4 & 18.8 & 1350.0 & 23.0 \\
 SIM~\citep{wieting-etal-2019-beyond} & 4/4 & 0.9 & 2.1 & 3.0 & 4.7 & 12.2 & 17.3 & 1358.2 & 11.0 \\
 DPR~\citep{karpukhin-etal-2020-dense} & 1/1 & 1.3 & 3.0 & 4.3 & 6.6 & 15.4 & 22.2 & 1205.3 & 25.5 \\
 DPR~\citep{karpukhin-etal-2020-dense} & 4/4 & 1.0 & 2.2 & 3.2 & 5.2 & 13.9 & 20.7 & 1208.1 & 22.5 \\
 c-REALM~\citep{lfqa21} & 1/1 & 1.6 & 3.5 & 4.8 & 7.1 & 15.9 & 21.7 & 1332.0 & 23.0\\
 c-REALM~\citep{lfqa21} & 4/4 & 0.9 & 2.1 & 3.3 & 5.0 & 12.9 & 18.8 & 1333.9 & 17.5 \\
 ColBERT~\citep{khattab2020colbert} & 1/1 & \textbf{2.9} & \textbf{6.0} & \textbf{7.8} & \textbf{11.0} & \textbf{21.4} & \textbf{27.9} & N/A\footnotemark  & \bf 38.8 \\
 ColBERT~\citep{khattab2020colbert} & 4/4 & 1.9 & 3.9 & 5.3 & 8.0 & 18.2 & 25.2 & N/A & 18.9 \\
\midrule  
\multicolumn{10}{c}{\textit{(trained on RELiC training set)}}\vspace{0.1cm}\\
\model & 0/1 & 3.4 & 7.1 & 9.3 & 12.6 & 24.1 & 31.3 & 1094.4 & 42.5 \\
   & 0/4 & 5.2 & 10.7 & 13.6 & 18.5 & 32.4 & 40.2 & 887.8 & 46.5 \\
 & 1/0 & 5.2 & 10.5 & 13.6 & 18.7 & 34.7 & 43.2 & 788.5 & \bf 67.5  \\
 & 4/0 & 6.8 & 14.4 & 19.3 & 25.7 & 43.9 & 52.8 & 538.3 & 65.5 \\
 & 1/1 & 7.8 & 15.1 & 19.3 & 25.7 & 43.3 & 52.0 & 558.0 & 67.0 \\
 & 4/4 & \textbf{9.4} & \textbf{18.3} & \textbf{24.0} & \textbf{32.4} & \textbf{51.3} & \textbf{60.8} & \textbf{377.3} & 65.0 \\
\midrule
Human domain experts & 4/4 & & & & & & & & \bf 93.5 \\
\bottomrule
\end{tabular}
\end{center}
\caption{Overall comparison of different systems and context sizes (\claim{L/R} indicates the number of sentences on the left and right side of the missing quote) on the test set of \name~using recall@$k$ metrics, normalized to a maximum score of 100. Our trained \model\ retriever significantly outperforms BM25 and all pretrained dense retrieval models. The average number of candidates per example is 4888. We report the accuracy of different systems\footnotemark on a proxy task that we administered to human domain experts, which shows that there is huge room for improvement.}
\label{tab:overall_compare}
\end{table*}

Having established that the examples in \name\ contain complex interplay between literary quotation and scholarly analysis, we now shift to measuring how well neural models can understand these interactions. In this section, we first formalize our evidence retrieval task, which provides \claim{the scholarly context \emph{without} the quotation} as input to a model, along with a set of candidate passages that come from the same book, and asks the model to retrieve the \litquote{ground-truth missing quotation} from the candidates. Then, we  describe standard information retrieval baselines as well as a RoBERTa-based ranking model that we implement to solve our task.


\subsection{Task formulation}
Formally, we represent a single window in \name\ from book $b$ as $(..., \claim{l_{-2}, l_{-1}}, \litquote{q_n}, \claim{r_1, r_2}, ...)$ where \litquote{$q_n$} is the quoted $n$-sentence long passage, and \claim{$l_i$} and \claim{$r_j$} correspond to individual sentences before and after the quotation in the scholarly article, respectively.  The window size on each side is bounded by hyperparameters $l_{max}$ and $r_{max}$, each of which can be up to 4 sentences. Given the \claim{$l_{-l_{max}:-1}$} and \claim{$r_{1: r_{max}}$} sentences surrounding the missing quotation, we ask models to identify the quoted passage \litquote{$q_n$} from the candidate set $C_{b,n}$, which consists of all $n$-sentence long passages in book $b$ (see Figure~\ref{fig:model_example}). This is a particularly challenging retrieval task because the candidates are part of the same overall narrative and thus mention the same overall set of entities (e.g., characters, locations) and other plot elements, which is a disadvantage for methods based on string overlap.

\paragraph{Evaluation:}
Models built for our task must produce a ranked list of candidates $C_{b,n}$ for each example. We evaluate these rankings using both recall@$k$ for $k=1,3,5,10,50,100$ and \emph{mean rank} of $q$ in the ranked list. Both types of metrics focus on the position of the ground-truth quotation \litquote{$q$} in the ranked list, and neither gives special treatment to candidates that overlap with  \litquote{$q$}. As such, recall@1 alone is overly strict when the quotation length $l>1$, which is why we show recall at multiple values of $k$. An additional motivation is that there may be multiple different candidates that fit a single context equally well. We also report accuracy on a proxy task with only three candidates, which allows us to compare with human performance as described in Section~\ref{sec:analysis}.

\footnotetext[8]{ColBERT does not provide a ranking for candidates outside the top 1000, so we cannot report mean rank.}
\footnotetext[9]{We do not report BM25's accuracy on the proxy task because its top-ranked quotes were used as candidates in the proxy task in addition to the ground-truth quotation.}

\subsection{Models}
\paragraph{Baselines:}  Our baselines include both standard term matching methods as well as pretrained dense retrievers. 
\textbf{BM25}~\citep{robertson1995okapi} is a bag-of-words method that is very effective for information retrieval.
We form queries by concatenating the \claim{left and right context} and use the implementation from the \href{https://github.com/dorianbrown/rank_bm25}{rank{\_}bm25} library\footnote{\url{https://github.com/dorianbrown/rank_bm25}, a library implementing many BM25-based algorithms.} to build a BM25 model for each unique candidate set $C_{b,n}$, tuning the free parameters as per~\citet{kamphuis2020bm25}.\footnote{We set $k_1$ = 0.5, $b$ = 0.9 after tuning on validation data.}

Meanwhile, our dense retrieval baselines are pretrained neural encoders that map \claim{queries} and \litquote{candidates} to vectors. We compute vector similarity scores (e.g., cosine similarity) between every query/candidate pair, which are used to rank candidates for every query and perform retrieval. 
We consider the following four pretrained dense retriever baselines in our work, which we deploy in a zero-shot manner (i.e., not fine-tuned on \name):

\begin{itemize}
    \item \textbf{DPR} (Dense Passage Retrieval) is a dense retrieval model from~\citet{karpukhin-etal-2020-dense} trained to retrieve relevant context paragraphs in open-domain question answering. We use the DPR context encoder\footnote{\url{https://huggingface.co/facebook/dpr-ctx_encoder-single-nq-base}} pretrained on Natural Questions~\citep{natural_questions} with dot product as a similarity function.
    \item \textbf{SIM} is a semantic similarity model from~\citet{wieting-etal-2019-beyond} that is effective on semantic textual similarity benchmarks~\citep{agirre-etal-2016-semeval}. SIM is trained on ParaNMT~\citep{wieting-gimpel-2018-paranmt}, a dataset containing 16.8M paraphrases; we follow the original implementation,\footnote{\url{https://github.com/jwieting/beyond-bleu}} and use cosine similarity as the similarity function. 
    \item \textbf{c-REALM} (contrastive Retrieval Augmented Language Model) is a dense retrieval model from~\citet{lfqa21} trained to retrieve relevant contexts in open-domain long-form question answering, and shown to be a better retriever than REALM~\citep{guu2020realm} on the ELI5 KILT benchmark~\citep{fan-etal-2019-eli5,petroni-etal-2021-kilt}. 
    \item \textbf{ColBERT} is a ranking model from~\citet{khattab2020colbert} that estimates the relevance between a query and a document using contextualized late interaction. It is trained on MS MARCO ranking data~\citep{nguyen2016ms}.
\end{itemize}

\paragraph{Training retrievers on \name~(\model):}
Both BM25 and the pretrained dense retriever baselines perform similarly poorly on \name\ (Table \ref{tab:overall_compare}). These methods are unable to capture more complex interactions within \name\ that do not exhibit extensive string overlap between quotation and context. As such, we also implement a strong neural retrieval model that is actually \emph{trained} on \name, using a similar setup to DPR and REALM. We first form a \claim{context string $c$} by concatenating a window of sentences on either side of the \litquote{quotation $q$} (replaced by a MASK token),
\begin{align*}
    \claim{c} = (\claim{l_{-l_{max}}, ..., l_{-1}}, \text{[MASK]}, \claim{r_1, ..., r_{r_{max}}})
\end{align*}
We train two encoder neural networks to project the \claim{literary context} and \litquote{quote} to fixed 768-$d$ vectors. Specifically, we project \claim{$c$} and \litquote{$q$} using \textbf{separate} encoder networks initialized with a pretrained RoBERTa-base model~\citep{liu2019roberta}. We use the \texttt{<s>} token of RoBERTa to obtain 768-$d$ vectors for the context and quotation, which we denote as \claim{$\mathbf{c}_i$} and \litquote{$\mathbf{q}_i$}.
To train this model, we use a contrastive objective~\citep{chen2020simple} that pushes the context vector \claim{$\mathbf{c}_i$} close to its quotation vector \litquote{$\mathbf{q}_i$}, but away from all other quotation vectors \litquote{$\mathbf{q}_j$} in the same minibatch (``in-batch negative sampling''):

\begin{align*}
    \text{loss} &= - \sum_{(\claim{c_i}, \litquote{q_i}) \in B} \log \frac{\exp \claim{\mathbf{c}_i} \cdot \litquote{\mathbf{q}_i}}{\sum_{\litquote{q_j} \in B} \exp \claim{\mathbf{c}_i} \cdot \litquote{\mathbf{q}_j}}
\end{align*}

\noindent where $B$ is a minibatch. Note that the size of the minibatch $|B|$ is an important hyperparameter since it determines the number of negative samples.\footnote{We set $|B|=100$, and train all models for 10 epochs on a single RTX8000 GPU with an initial learning rate of 1e-5 using the Adam optimizer \cite{adam_kingma}, early stopping on validation loss. Models typically took 4 hours to complete 10 epochs. Our implementation uses the HuggingFace \texttt{transformers} library \cite{wolf-etal-2020-transformers}. The total number of model parameters is 249M.} All elements of the minibatch are context/quotation pairs sampled from the same book. During inference, we rank all \litquote{quotation candidate vectors} by their dot product with the \claim{context vector}.

\subsection{Results}
\label{sec:results}

We report results from the baselines and our \model\ model in Table~\ref{tab:overall_compare} with varying context sizes where \claim{$L/R$} refers to  $L$ preceding context sentences and $R$ subsequent context sentences. While all models substantially outperform random candidate selection, all pretrained neural dense retrievers perform similarly to BM25, with ColBERT being the best pretrained neural retriever (2.9 recall@1). This result indicates that matching based on string overlap or semantic similarity is not enough to solve \name, and even powerful neural retrievers struggle on this benchmark. Training on \name\ is crucial: our best-performing \model\ model performs 7x better than BM25 (9.4 vs 1.3 recall@1).

\paragraph{Context size and location matters for model performance:} Table~\ref{tab:overall_compare} shows that \model\ effectively utilizes longer context --- feeding only one sentence on each side of the quotation (1/1) is not as effective  as a longer context (4/4) of four sentences on each side (7.8 vs 9.4 recall@1). However, the longer contexts hurt performance for pretrained dense retrievers in the zero-shot setting (1.6 vs 0.9 recall@1 for c-REALM), perhaps because context further away from the quotation is less likely to be helpful. Finally, we observe that \model\ performance is strictly better (5.2 vs 6.8 recall@1) when the model is given only preceding context (4/0 or  1/0) compared to when the model is given only subsequent context (0/4 or 0/1).


\paragraph{Dense vs. sparse retrievers:}
As expected, BM25 retrieves the correct quotation when there is significant string overlap between the quotation and context, as in the following example from \textit{The Great Gatsby}, in which the terms \textit{sky}, \textit{bloom}, \textit{Mrs. McKee}, \textit{voice}, \textit{call}, and \textit{back} appear in both places:
\begin{quote}
\small
   \claim{Yet his analogy also implicitly unites the two women. Myrtle's expansion and revolution in the smoky air are also outgrowths of her surreal attributes, stemming from her residency in the Valley of Ashes.} \litquote{\textbf{The late afternoon sky bloomed in the window for a moment like the blue honey of the Mediterranean-then the shrill voice of Mrs. McKee called me back into the room.}} \claim{The objective talk of Monte Carlo and Marseille has made Nick daydream. In Chapter I Daisy and the rooms had bloomed for him, with him, and now the sky blooms. The fact that Mrs. McKee's voice ``calls him back'' clearly reveals the subjective daydreamy nature of this statement.}
\end{quote}

However, this behavior is undesirable for most examples in \name, since string overlap is generally not predictive of the relationship between quotations and claims. The top row of Table~\ref{tab:dense_vs_bm25} contains one such example, where \model\ correctly chooses the missing quotation while BM25 is misled by string overlap. 


\section{Human performance and analysis}
\label{sec:analysis}
How well do humans actually perform on \name? 
To compare the performance of our dense retriever to that of humans, we hired six domain experts with at least undergraduate-level degrees in English literature from the Upwork\footnote{\url{https://upwork.com}} freelancing platform. Because providing thousands of candidates to a human evaluator is infeasible, we instead measure human performance on a simplified proxy task: we provide our evaluators with four sentences on either side of a missing quotation from \emph{Pride and Prejudice}\footnote{We decided to keep our proxy task restricted to the most well-known book in our test set because of the ease with which we could find highly-qualified workers who self-reported that they had read (and often even re-read) \emph{Pride and Prejudice}.} and ask them to select one of only three candidates to fill in the blank. We obtain human judgments both to measure a \emph{human upper bound} on this proxy task  as well as to evaluate whether humans struggle with examples that fool our model.


\paragraph{Human upper bound:}
First, to measure a human upper bound on this proxy task, we chose 200 test set examples from \emph{Pride and Prejudice} and formed a candidate pool for each by including BM25's top two ranked answers along with the ground-truth \litquote{quotation} for the single sentence case. As the task is trivial to solve with random candidates, we decided to use a model to select harder negatives, and we chose BM25 to see if humans would be distracted by high string overlap in the negatives. Each of the 200 examples was separately annotated by three experts,  and they were paid \$100 for annotating 100 examples. The last column of Table~\ref{tab:overall_compare} compares all of our baselines along with \model\ against human domain experts on this proxy task. Humans substantially outperform all models on the task, with at least two of the three domain experts selecting the correct quote 93.5\% of the time; meanwhile, the highest score for \model\ is 67.5\%, which indicates huge room for improvement. Interestingly, all of the zero-shot dense retrievers except ColBERT 1/1 underperform random selection on this task; we theorize that this is because all of these retrievers are misled by the high string overlap of the negative BM25-selected examples. Table~\ref{tab:inter-annotator-agreement} confirms substantial agreement among our annotators.

\begin{table}[h!]
\small
\begin{center}
\begin{tabular}{lrrr}  
\toprule
 & Fleiss $\kappa$ ($\uparrow$) & all agree ($\uparrow$) & none agree ($\downarrow$) \\
 \midrule
 Random & 0.00 & 11.1\% & 22.2\% \\
 Humans & 0.68 & 68.5\% & 0.5\% \\
 \bottomrule
\end{tabular}
\end{center}
\caption{Inter-annotator agreement of our three human annotators compared to a random annotation. In our 3-way classification task, all three annotators chose the same option 68.5\% of the time, while they each chose a different option in just 0.5\% of instances. Our annotators also show substantial agreement in terms of Fleiss Kappa~\citep{fleiss1971measuring}.\footnotemark[17]}
\label{tab:inter-annotator-agreement}
\end{table}

\footnotetext[17]{In our proxy task each instance has a different set of \litquote{candidate quotations}, which we randomly shuffle before showing annotators. Since our task is not strictly categorical, while computing Fleiss Kappa we define ``category'' as the option number shown to annotators. We believe this definition is closest to the free-marginal nature of our task~\citep{randolph2005free}.}

\begin{table*}[ht]
  \small
  \centering
  \footnotesize
  \begin{tabular}{m{3.5cm} m{3.5cm} m{3.5cm} m{3.5cm}}
    \toprule
    \bf Surrounding context & \bf Correct candidate & \bf Incorrect candidate & \bf Analysis\\
    \midrule
    \claim{She is caught up for a moment or two in a fantasy of possession:} \litquote{\textbf{[masked quote]}} \claim{The thought that she would not have been allowed to invite the Gardiners is a lucky recollection it save[s] her from something like regret.} \cite{paris_character_1978} & [\emph{\model}]: \litquote{``And of this place,'' thought she, ``I might have been mistress! With these rooms I might now have been familiarly acquainted!''} 
     & [\emph{BM25}]: ``I should not have been allowed to invite them.'' This was a lucky recollection-it saved her from something very like regret. & \model\ correctly retrieves the quotation that shows the ``fantasy of possession,'' while BM25 retrieves a quote that is paraphrased in the surrounding context. \\
     \midrule
     \claim{It is delicious from the opening sentence:} \litquote{\textbf{[masked quote]}} \claim{Mr. Bingley, with his four or five thousand a year, had settled at Netherfield Park.} \cite{masefield_women_1967} & [\emph{Human}]: \litquote{It is a truth universally acknowledged, that a single man in possession of a good fortune, must be in want of a wife.} & [\emph{dense-RELiC}]: ``My dear Mr. Bennet,'' said his lady to him one day, ``have you heard that Netherfield Park is let at last?'' & Human readers can immediately identify the first sentence of \textit{Pride and Prejudice}, while \model\ lacks this world knowledge.\\
     \midrule
     \claim{Sometimes we hear Mrs Bennet's idea of marriage as a market in a single word:} \litquote{\textbf{[masked quote]}} \claim{Her stupidity about other people shows in all her dealings with her family...} \cite{mcewan_style_1986} & [\emph{Human}]: \litquote{``I do not blame Jane,'' she continued, ``for Jane would have got Mr. Bingley if she could.''} & [\emph{dense-RELiC}]: You must and shall be married by a special licence. & Human readers understood the uncommon usage of ``got'' to convey a transaction.\\

    \bottomrule
  \end{tabular}
  \caption{Examples that show failure cases of BM25 (top row) and our \model\ retriever (bottom two rows) from our proxy task on \emph{Pride and Prejudice}. BM25 is easily misled by string overlap, while \model\ lacks world knowledge (e.g., knowing the famous first sentence) and complex linguistic understanding (e.g., the relationship between \claim{marriage as a market} and \litquote{got}) that humans can easily rely on to disambiguate the correct quotation.}
  \label{tab:dense_vs_bm25}
\end{table*}

\paragraph{Human error analysis of \model:}
To evaluate the shortcomings of our \model\ retriever, we also administered a version of the proxy task where the candidate pool included the ground-truth quotation along with \model's two top-ranked candidates, where for all examples the model ranked the ground-truth outside of the top 1000 candidates. Three domain experts attempted 100 of these examples and achieved an accuracy of 94\%, demonstrating that humans can easily disambiguate cases on which our model fails, though we note our model's poorer performance when retrieving a single sentence (as in the proxy task) versus multiple sentences (\ref{tab:prompt_with_context}). The bottom two rows of Table~\ref{tab:dense_vs_bm25} contain instances in which all human annotators agreed on the correct candidate but \model\ failed to rank it in the top 1000. In one, all human annotators immediately recognized the opening line of \textit{Pride and Prejudice}, one of the most famous in English literature. In the other, the claim mentions that the interpretation hinges on a single word's (``got'') connotation of ``a market,'' which humans understood.

\begin{table*}[t]
\footnotesize
  \centering
  \begin{tabular}{m{16cm}}
    \toprule
    From \textit{Frankenstein}, given \claim{``Victor does not consider the consequences of his actions:''} our model's top-ranked single sentence candidates are:\\ \vspace{0.1cm}
    1. \litquote{It is even possible that the train of my ideas would never have received the fatal impulse that led to my ruin.} \\
    2. \litquote{The threat I had heard weighed on my thoughts, but I did not reflect that a voluntary act of mine could avert it.} \\
    3. \litquote{Now my desires were complied with, and it would, indeed, have been folly to repent.}\vspace{0.1cm} \\
    \midrule
    From \textit{The Great Gatsby}, given \claim{``A symbol of Gatsby's lifestyle:''} our model's top-ranked single sentence candidates are:\\ \vspace{0.1cm}
    1. \litquote{His movements-he was on foot all the time-were afterward traced to Port Roosevelt and then to Gad's Hill where he bought a sandwich that he didn't eat and a cup of coffee.} \\
    2. \litquote{Every Friday five crates of oranges and lemons arrived from a fruiterer in New York-every Monday these same oranges and lemons left his back door in a pyramid of pulpless halves.} \\
    3. \litquote{On week-ends his Rolls-Royce became an omnibus, bearing parties to and from the city, between nine in the morning and long past midnight, while his station wagon scampered like a brisk yellow bug to meet all trains.}\vspace{0.1cm}\\
    \midrule
    From \textit{Pride and Prejudice}, given \claim{``Elizabeth displays frustration towards her mother:''} our model's top-ranked 2-sentence candidates are:\\ \vspace{0.1cm}
    1. \litquote{Oh, that my dear mother had more command over herself!  She can have no idea of the pain she gives me by her continual reflections on him.} \\
    2. \litquote{My mother means well;  but she does not know, no one can know, how much I suffer from what she says.} \\
    3. \litquote{with tears and lamentations of regret, invectives against the villainous conduct of Wickham, and complaints of her own sufferings and ill-usage;  blaming everybody but the person to whose ill-judging indulgence the errors of her daughter must principally be owing.} \\
    \midrule
    From \textit{Brave New World}, given \claim{``Children are indoctrinated while sleeping and taught hypnopaedic phrases, such as''}, our model's top-ranked single sentence candidates are:\\ \vspace{0.1cm}
    1. \litquote{The principle of sleep-teaching, or hypnopædia, had been discovered.} \\
    2. \litquote{Roses and electric shocks, the khaki of Deltas and a whiff of asafoetida-wedded indissolubly before the child can speak.} \\
    3. \litquote{Told them of the growing embryo on its bed of peritoneum.}\\
  \bottomrule
  \end{tabular}
  \caption{Given a novel and a short out-of-distribution \claim{prompt}, this table shows the top 3 \litquote{quotations} from the novel that dense-RELiC returns as evidence. The relevance of many of the returned quotations, even without string overlap between the prompt and candidates, indicates the model is learning some non-trivial relationships that could have potential impact for building tools that support humanities research. However, it is not perfect, as shown in the final example where none of the retrieved quotations is actually an instance of a hypnopaedic phrase.}
  \label{tab:prompt}
\end{table*}

\paragraph{Issuing out-of-distribution queries to the retriever:}
Does our \model\ model have potential to support humanities scholars in their evidence-gathering process? 
Inspired by prompt-based learning, we manually craft simple yet out-of-distribution prompts and queried our \model\ retriever trained with 1 sentence of left context and no right context. A qualitative inspection of the top-ranked quotations in response to these prompts (Table \ref{tab:prompt}) reveals that the retriever is able to obtain evidence for  distinct character traits, such as the ignorance of the titular character in \textit{Frankenstein} or Gatsby's wealthy lifestyle in \textit{The Great Gatsby}. More impressively, when queried for an example from \textit{Pride and Prejudice} of the main character, Elizabeth, demonstrating frustration towards her mother, the retriever returns relevant excerpts in the first-person that do not mention Elizabeth, and the top-ranked quotations have little to no string overlap with the prompts. 

\paragraph{Limitations:} While these results show \model's potential to assist research in the humanities, the model suffers from the limited expressivity of its candidate quotation embeddings \litquote{$\mathbf{q}_i$}, and addressing this problem is an important direction for future work. The quotation embeddings do not incorporate any broader context from the narrative, which prevents resolving coreferences to pronominal character mentions and understanding other important discourse phenomena.  For example, Table \ref{tab:prompt_with_context} shows that \model\ 's top two 1-sentence candidates for the above \emph{Pride and Prejudice} example are not appropriate evidence for the literary claim; the increased relevancy of the 2-sentence candidates (Table~\ref{tab:prompt}, third row) over the 1-sentence candidates suggests that \model\ may benefit from more contextualized quotation embeddings. Furthermore, \model\ struggles with retrieving concepts unique to a text, such as the ``hypnopaedic phrases'' strewn throughout \textit{Brave New World} (Table~\ref{tab:prompt}, bottom).
 


\section{Related Work}
\label{sec:related_work}

\paragraph{Datasets for literary analysis:}
Our work relates to previous efforts to apply NLP to literary datasets such as LitBank~\citep{bamman2019annotated,sims2019literary}, an annotated dataset of 100 works of fiction with annotations of entities, events, coreferences, and quotations. \citet{papay-pado-2020-riqua} introduced RiQuA, an annotated dataset of quotations in English literary text for studying dialogue structure, while~\citet{chaturvedi2016modeling} and ~\citet{rmn2016} characterize character relationships in novels. Our work also relates to quotability identification~\citep{maclaughlin2021content}, which focuses on ranking passages in a literary work by how often they are quoted in a larger collection. Unlike \name, however, these datasets do not contain literary analysis about the works. 

\paragraph{Retrieving cited material:}
Citation retrieval closely relates to \name\ and has a long history of research, mostly on scientific papers:
\citet{oconnor_citing_1982} formulated the task of document retrieval using ``citing statements'', which  \citet{liu2014literature} revisit to create a reference retrieval tool that recommends references given context. \citet{bertin2016linguistic} examine the rhetorical structure of citation contexts. Perhaps closest to \name\ is the work of~\citet{grav_harnessing_2019}, which concentrates on the quotation of secondary sources in other secondary sources, unlike our focus on quotation from primary sources. Finally, as described in more detail in Section~\ref{sec:compare-other-datasets} and Appendix~\ref{tab:compare-beir}, \name\ differs significantly from existing NLP and IR retrieval datasets in domain, linguistic complexity, and query length.



\section{Conclusion}
\label{sec:conclusion}
In this work, we introduce the task of \textit{literary evidence retrieval} and an accompanying dataset, \name. We find that \litquote{direct quotation} of primary sources in literary analysis is most commonly used as evidence for \claim{literary claims or arguments}. We train a dense retriever model for our task; while it significantly outperforms baselines, human performance indicates a large room for improvement. Important future directions include (1) building better models of \emph{primary sources} that integrate narrative and discourse structure into the candidate representations instead of computing them out-of-context, and (2) integrating \name\ models into real tools that can benefit humanities researchers.
\section*{Acknowledgements}
First and foremost, we would like to thank the HathiTrust Research Center staff (especially Ryan Dubnicek) for their extensive feedback throughout our project.
We are also grateful to Naveen Jafer Nizar for his help in cleaning the dataset, Vishal Kalakonnavar for his help with the project webpage, Marzena Karpinska for her guidance on computing inter-annotator agreement, and the UMass NLP community for their insights and discussions during this project. KT and MI are supported by awards IIS-1955567 and IIS-2046248	from the National Science Foundation (NSF). KK is supported by the Google PhD Fellowship awarded in 2021.
\section*{Ethical Considerations}

We acknowledge that the group of authors from whom we selected primary sources lacks diversity because we selected from among digitized, public domain sources in the Western literary canon, which is heavily biased towards white, male writers. We made this choice because there are relatively few primary sources in the public domain that are written by minority authors and also have substantial amounts of literary analysis written about them. We hope that our data collection approach will be followed by those with access to copyrighted texts in an effort to collect a more diverse dataset. The experiments involving humans were reviewed by the UMass Amherst IRB with a status of Exempt.

\bibliography{bib/journal-full,bib/anthology,bib/custom}
\bibliographystyle{bib/acl_natbib}

\clearpage
\setcounter{table}{0}
\renewcommand{\thetable}{A\arabic{table}}
\appendix
\section*{Appendices for ``\name: Retrieving Evidence from Literature in Context''}
\label{sec:appendix}

\section{Dataset Collection \& Statistics}
\label{appendix:data}

\paragraph{Filtering secondary sources:} The HathiTrust is not exclusively a repository of literary analysis, and we observe that many matching quotes come from different editions of a primary source, writing manuals, and even advertisements. Because we are seeking only scholarly work that directly analyzes the quoted sentences, we performed a combination of manual and automatic filtering to remove such extraneous matches. For each primary source, we first aggregate all secondary sources matches by the their unique HathiTrust-assigned identifier. From manual inspection of the secondary source titles, most sources that quote a particular literary work only once or twice are not likely to be literary scholarship, while sources with hundreds of matches are almost always a different edition of the primary source itself. For each primary source, we create upper and lower thresholds for number of matches, discarding sources that fall outside of these bounds. Additionally, we discard secondary sources whose titles contain the words ``dictionary'', ``anthology'', ``encyclopedia,'' and others that indicate that a secondary source is not literary scholarship.

\paragraph{Preprocessing:}
After the above filtering, we identified and removed all non-English secondary sources using langid,\footnote{\url{https://github.com/saffsd/langid.py}} a Python tool for language identification. Next, because the secondary source texts in the HathiTrust are digitized via OCR, various artifacts appear throughout the pages we download. Some of these, such as citations that include the page number of primary source quotes, allow models trained on our task to ``cheat'' to identify the proper quote (see Table \ref{tab:clean_text}), necessitating their removal. Using a pattern-matching approach, we eliminate the most pervasive: in-line citations, headers, footers, and word breaks. Finally, we apply sentence tokenization in order to standardize the length of preceding and subsequent context windows for the final dataset. Specifically, we feed the preprocessed text through spaCy's\footnote{\url{https://spacy.io/}} dependency parser-based sentence segmenter on the cleaned text. The default segmenter in spaCy is modified to use ellipses, colons, and semicolons as custom sentence boundaries, based on the observation that literary scholars often only quote part of what would typically be defined as a sentence (Table \ref{tab:partial_sent_quote}).

\begin{table}[h]
\footnotesize
\begin{center}
\begin{tabular}{ m{7cm}}  
\toprule
 \textit{Raw text from HathiTrust:} \\
 \midrule
 The prejudice in these same eyes, however, keeps them ``less clear-sighted'' \textbf{(p. 149)} to Bingley's feelings for Jane and totally closed to the real \textbf{worth- lessness} of Wickham and worth of Darcy. When Jane's letter reporting \textbf{196 Mark M. Hennelly, Jr.} Lydia's disappearance with Wickham confirms Darcy's earlier indictment of him, though, Elizabeth's ``eyes were opened to his real character'' \textbf{(p. 277)}.\\
 \bottomrule
\end{tabular}
\end{center}
\caption{An analysis of Jane Austen's \textit{Pride and Prejudice} from \citet{hennelly_eyes_1983} that contains artifacts (bold) such as citations and page numbers that we remove during preprocessing.}
\label{tab:clean_text}
\end{table}

\begin{table}[h]
\footnotesize
\begin{center}
\begin{tabular}{ m{7cm}}  
\toprule
 \textit{Quoted span in context of literary analysis:} \\
 Edna tries to discuss this issue of possession versus self-possession with Madame Ratignolle but to no avail; `\textbf{the two women did not appear to understand each other or to be talking the same language.}' Madame Ratignolle cannot comprehend that there might be something more that a mother could sacrifice for her children beyond her life...\\
 \midrule\
 \textit{Quote in original context from The Awakening:}\\
 Edna had once told Madame Ratignolle that she would never sacrifice herself for her children, or for any one. Then had followed a rather heated argument; \textbf{the two women did not appear to understand each other or to be talking the same language.} Edna tried to appease her friend, to explain.\\
 \bottomrule
\end{tabular}
\end{center}
\caption{An analysis of Kate Chopin's \textit{The Awakening} from \citet{madsen_feminist_2000} that quotes part of a sentence (following a semi-colon) from the primary source. We detect such partial matches during preprocessing.}
\label{tab:partial_sent_quote}
\end{table}

\paragraph{Identifying quoted sentences:} As previously mentioned, HathiTrust does not provide the exact indices corresponding to the primary source quote. As such, we identify which secondary source sentences (from the output of the sentence tokenizer) include quotes from primary source works using RapidFuzz, \footnote{\url{https://github.com/maxbachmann/RapidFuzz}} a fuzzy string match library, with the QRatio metric and a score threshold of 80.0. Fuzzy match is essential for detecting quotes with OCR mistakes or with author modifications; in Appendix Table~\ref{tab:block_fuzzy_example}, for instance, the author adds clarification [the natives] and omits ``he would say'' when citing two sentences from Joseph Conrad's \textit{Heart of Darkness}. Once a fuzzy match is identified in a secondary source document, we replace it with its corresponding primary source sentence.

\begin{table}[h]
\footnotesize
\begin{center}
\begin{tabular}{ m{7cm}}  
\toprule
     \textit{Secondary source material:} \\
    Kurtz's credo, like his royal employer's, was a simple one.\\
    1. ``You show them [the natives] you have in you something that is really profitable, and then there will be no limits to the recognition of your ability.\\
    2. Of course you must take care of the motives---right motives---always.''\\
    Kurtz dies screaming: "The Horror! The Horror!" Leopold, so far as one knows, died more peacefully \cite{legum_congo_1972}.\\
    \midrule
    \textit{Window in \name\ with standardized quote:}\\
    Kurtz's credo, like his royal employer's, was a simple one. \textbf{`You show them you have in you something that is really profitable, and then there will be no limits to the recognition of your ability,' he would say. `Of course you must take care of the motives---right motives---always.'} Kurtz dies screaming: "The Horror! The Horror!" Leopold, so far as one knows, died more peacefully.\\
 \bottomrule
 
\end{tabular}

\end{center}
\caption{This example demonstrates the necessity of fuzzy match and block quote identification. Consecutive sentences are quoted and one is slightly modified from its original form in the primary source.}
\label{tab:block_fuzzy_example}
\end{table}

\paragraph{Identifying block quotes:} 
While we query HathiTrust at a sentence level, many of the returned results are actually \emph{block quotes} in which multiple contiguous sentences from the primary source are quoted. Correct identification of these block quotes is integral to the quality of our dataset and formulated task: if the preceding or subsequent context contains part of the quoted span, our evidence retrieval task becomes trivial because part of the answer exists in the input. In our approach, if the fuzzy match yields consecutive matches in secondary source documents for sentences that also appear consecutively in the primary source, we concatenate them together and consider them a single block quote.

\paragraph{Handling ellipses:}
 One prevalent technique for direct quotation in literary analysis is the use of ellipses to condense primary source material. As our fuzzy match method still falls short in detecting block quotes that contain ellipses, we implement an additional method for insuring that block quotes are properly delineated. Once the fuzzy match approach fails to identify any more consecutively quoted sentences in a secondary source, we continue to search for matches adjacent to the block quote using the Longest Common Substring (LCS) metric. If a block-quote-adjacent sentence in the secondary source shares an LCS of 15 or more characters with the block-quote-adjacent sentence in the primary source, this is considered a match and concatenated with the block quote (see Appendix~\ref{appendix:lcs} for an example).

\subsection{LCS example}
\label{appendix:lcs}
For example, in \citet{dabydeen_revelation_1985}, Kenneth Parker cites a passage from Joseph Conrad's \textit{Heart of Darkness}: ``The narrator, Marlow, informs us, approvingly:...\textbf{I met a white man, in such an unexpected elegance of get-up that in the first moment I took him for a sort of vision.} I saw a high starched collar, white cuffs, a light alpaca jacket, snowy trousers, a clean necktie, and varnished boots.'' Fuzzy match alone is insufficient for detecting the first sentence in this block quote that contains an ellipse in place of primary source text. With our LCS approach, we are able to replace the first sentence of block quote above with ``\textbf{When near the buildings I met a white man, in such an unexpected elegance of get-up that in the first moment I took him for a sort of vision.}'' 

\subsection{Noise when standardizing  quotes:}
\label{appendix:standardization}
In a small number of cases, our quote standardization process removes important context. For example, the analysis of \citet{maes-jelinek_criticism_1970} quotes a sentence from D.H. Lawrence's \emph{The Rainbow} as ``As to Will, \textbf{his intimate life was so violently active, that it set another man free in him.}''. After standardization, the example in our dataset becomes ``\textbf{His intimate life was so violently active, that it set another man free in him.}'', dropping the critical ``As to Will'' necessary for the integration of the quote in the surrounding analysis.

\begin{table}[t]
\footnotesize
\begin{center}
\begin{tabular}{ m{7cm}}  
\toprule
    \textbf{Window of secondary source analysis:}\\
    \midrule
    For example, Elizabeth's anger with herself, after reading Darcy's letter, is couched largely in the vocabulary of rectifiable intellectual error"blind, partial, prejudiced, absurd, and the like-rather than in the relentless, coercive vocabulary of moral contrition. Her discomfiture, though profound, has a Greek ring to it: \textbf{Till this moment I never knew myself.} Heuristically, the distinction between moral and other spheres of value throws light also on other Austen novels that we can only glance at here \cite{wilkie_jane_1992}.\\
    \midrule
    \textbf{Best model's top ranked candidate:}\\
    \midrule
    that loss of virtue in a female is irretrievable;\\
    \midrule
    \textbf{Best model's second ranked candidate}\\
    \midrule
    but when she considered how unjustly she had condemned and upbraided him, her anger was turned against herself;\\
 \bottomrule
\end{tabular}
\end{center}
\caption{The model ranked the correct quote outside of the top ten percent of 5,278 candidates, but all 3 domain experts selected the model's second ranked candidate over the ground-truth quote.}
\label{tab:experts_and_model_wrong}
\end{table}

\paragraph{Model-predicted quotes are sometimes as valid as the gold quote:}
Human raters also identify cases in which multiple quotes appear to be appropriate evidence for a literary claim, which illustrate the model's potential in helping humanities scholars find evidence. In Table \ref{tab:experts_and_model_wrong}, both model and experts failed to identify the correct quote that both depicts Elizabeth's 
``discomfiture'' and has a ``Greek ring to it:'' ``Till this moment I never knew myself.'' However, the experts all selected the model's second ranked choice which mentions Elizabeth's ``anger'' at ``herself.'' This quote also shows Elizabeth's displeasure while referring to the Greek idea of self.

\subsection{More dataset statistics}
\label{sec:more-data-stats}

\begin{figure}[h]
\centering
\includegraphics[width=5.5cm]{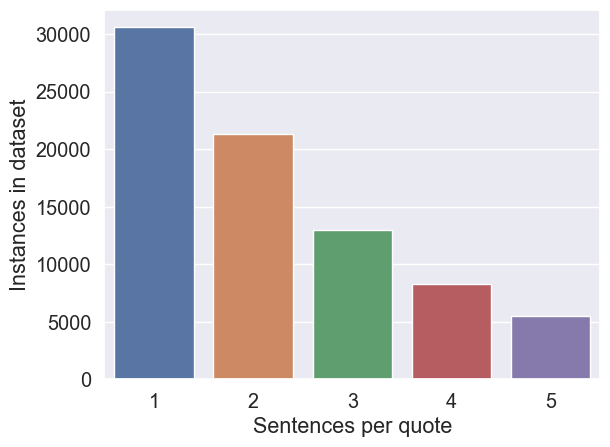}
\caption{Distribution of \name\ quote lengths.}
\label{fig:distribution}
\end{figure}

Each primary source has relevant windows from an average of 112 unique secondary sources, and an average of 16.35\% of the sentences in each primary source are quoted in secondary sources.  On average, each primary source has 995 corresponding windows in our dataset, and each secondary source produced an average of 9 windows.
Figure \ref{fig:distribution} shows the distribution of quote lengths in \name, suggesting that successful models will have to learn to understand both single-sentence and block quotes in context. 

\section{Best Model Detailed Results}
\paragraph{Candidate length does not significantly affect model performance:}
We observe in Table \ref{tab:best_model} that the length of the \litquote{ground-truth quote} and the candidates does not significantly impact model performance --- for a fixed $k$, model performance is within 10\% for any candidate length. Model performance is slightly worse for longer candidates of length 4 or 5, and for the shortest single sentence contexts (possibly due to under-specification). 



\begin{table*}[t]
\footnotesize
  \centering
  \begin{tabular}{m{16cm}}
    \toprule
    From \textit{Pride and Prejudice}, given \claim{''Elizabeth displays frustration towards her mother:''} our model's top-ranked, 1-sentence candidates are:\\ \vspace{0.1cm}
    1. \litquote{Elizabeth was again deep in thought, and after a time exclaimed, "To treat in such a manner the godson, the friend, the favourite of his father!"} \\
    2. \litquote{Far be it from me," he presently continued, in a voice that marked his displeasure, "to resent the behaviour of your daughter.} \\
    3. \litquote{Her mother's ungraciousness, made the sense of what they owed him more painful to Elizabeth's mind;}\\
  \bottomrule
  \end{tabular}
  \caption{When querying the model using out-of-distribution \claim{prompts}, number of sentences of the desired candidates can be specified. This table shows the top 3 \litquote{quotations} from the \textit{Pride and Prejudice} that dense-RELiC returns as evidence for single-sentence candidates. The suitability of the 2-sentence candidates (show in Table \ref{tab:prompt}) over the single-sentence candidates suggests that contextualizing the \litquote{quotation} embeddings will improve model performance.}
  \label{tab:prompt_with_context}
\end{table*}

\begin{table*}[t!]
    \small
    \resizebox{\textwidth}{!}{\begin{tabular}{ l | l | l | c | c | c | c | c c c | c c }
        \toprule
         \multicolumn{1}{l}{\textbf{Split} ($\rightarrow$)} &
         \multicolumn{4}{c}{} &
         \multicolumn{1}{c}{\textbf{Train}}    &
         \multicolumn{1}{c}{\textbf{Dev}}    &
         \multicolumn{3}{c}{\textbf{Test}}   &
         \multicolumn{2}{c}{\textbf{Avg.~Word Lengths}} \\
         \cmidrule(lr){6-6}
         \cmidrule(lr){7-7}
         \cmidrule(lr){8-10}
         \cmidrule(lr){11-12}
           \textbf{Task ($\downarrow$)} &\textbf{Domain ($\downarrow$)} & \textbf{Dataset ($\downarrow$)} & \textbf{Title} & \textbf{Relevancy} & \textbf{\#Pairs} & \textbf{\#Query} & \textbf{\#Query} & \textbf{\#Corpus} & \textbf{Avg. D~/~Q } & \textbf{Query} & \textbf{Document} \\
         \midrule
    Passage-Retrieval    & Misc. & MS MARCO \cite{nguyen2016ms} & \xmark & Binary  & 532,761 &   ----  &   6,980   &   8,841,823      & 1.1 & 5.96  & 55.98  \\ \midrule[0.05pt] \midrule[0.05pt]
    Bio-Medical          & Bio-Medical & TREC-COVID \cite{voorhees2021trec}  & \cmark & 3-level &   ----    &   ----  & 50     & 171,332   & 493.5& 10.60 & 160.77 \\
    Information          & Bio-Medical & NFCorpus \cite{boteva2016}      & \cmark & 3-level & 110,575 &  324  & 323    & 3,633     & 38.2 & 3.30  & 232.26 \\
    Retrieval (IR)       & Bio-Medical & BioASQ \cite{tsatsaronis2015overview}       & \cmark & Binary  & 32,916 & ---- & 500    & 14,914,602& 4.7  & 8.05  & 202.61 \\ \midrule
    Question             & Wikipedia  & NQ  \cite{natural_questions}           & \cmark & Binary  & 132,803  &   ----  & 3,452 & 2,681,468 & 1.2  & 9.16  & 78.88  \\
    Answering       & Wikipedia  & HotpotQA \cite{yang-etal-2018-hotpotqa}     & \cmark & Binary  & 170,000 & 5,447 & 7,405  & 5,233,329 & 2.0  & 17.61 & 46.30  \\
     (QA)           &Finance& FiQA-2018 \cite{10.1145/3184558.3192301}  & \xmark & Binary  & 14,166  &  500  & 648    & 57,638    & 2.6  & 10.77 & 132.32 \\ \midrule
    Tweet-Retrieval      &Twitter& Signal-1M (RT)  \cite{Signal1MRelatedTweetsRetrieval2018}    & \xmark & 3-level &   ----    &   ----  & 97     & 2,866,316 & 19.6 & 9.30  & 13.93  \\ \midrule
    News      &News& TREC-NEWS  \cite{soboroff2018trec}    & \cmark & 5-level &   ----    &   ----  & 57     & 594,977 & 19.6 & 11.14  & 634.79  \\
    Retrieval      &News& Robust04 \cite{96071}    & \xmark & 3-level &   ----    &   ----  & 249   & 528,155 & 69.9 & 15.27  & 466.40  \\ \midrule
    Argument       & Misc. & ArguAna  \cite{wachsmuth:2018a}    & \cmark & Binary  &   ----    &   ----  & 1,406  & 8,674     & 1.0  & \textbf{192.98} & 166.80 \\
    Retrieval   & Misc. & Touch\'e-2020 \cite{stein:2020v} & \cmark & 3-level &   ----    &   ----  & 49     & 382,545   & 19.0 & 6.55  & 292.37 \\ \midrule
    Duplicate-Question   &StackEx.& CQADupStack \cite{hoogeveen2015cqadupstack}  & \cmark & Binary  &   ----    &   ----  & 13,145 & 457,199   & 1.4  & 8.59  & 129.09 \\
    Retrieval            & Quora &  Quora        & \xmark & Binary  &   ----    & 5,000 & 10,000 & 522,931   & 1.6  & 9.53  & 11.44  \\ \midrule
    Entity-Retrieval     & Wikipedia  &  DBPedia  \cite{Hasibi:2017:DVT}     & \cmark & 3-level &   ----    &   67  & 400    & 4,635,922 & 38.2 & 5.39  & 49.68  \\ \midrule
    Citation-Prediction  & Scientific&  SCIDOCS  \cite{cohan-etal-2020-specter}     & \cmark & Binary  &   ----    &   ----  & 1,000  & 25,657    & 4.9  & 9.38  & 176.19 \\ \midrule
                         & Wikipedia  &  FEVER \cite{thorne-etal-2018-fever}       & \cmark & Binary  & 140,085 & 6,666 & 6,666  & 5,416,568 & 1.2  & 8.13  & 84.76  \\ 
    Fact Checking        & Wikipedia  & Climate-FEVER \cite{diggelmann2020climatefever} & \cmark & Binary  &   ----    &   ----  & 1,535  & 5,416,593 & 3.0  & 20.13 & 84.76  \\
                         & Scientific & SciFact  \cite{wadden-etal-2020-fact}     & \cmark & Binary  &   920      &   ----  &  300   & 5,183     & 1.1  & 12.37 & 213.63  \\
    \midrule
    \midrule
    \textbf{Literary evidence retrieval} & \textbf{Literature} & \name\ (this work) & \xmark & Binary & 71395 & 9036 & 9034 & 5041 & 1.0 & \textbf{154.1} & 45.5 \\
    \bottomrule
    \end{tabular}}
    \caption{A comparison between datasets in the BEIR benchmark and our \name\ dataset. Ours is the first retrieval dataset in the literary domain, formulating a new task of literary evidence retrieval.}
    \label{tab:compare-beir}
\end{table*}

\begin{table*}[h]
\footnotesize
\centering
\begin{tabular}{lllll}
\toprule
 & & \textbf{Training Set} & & \\
\midrule
\textbf{Year} & \textbf{Title}                                   & \textbf{Author (Translator)}                 & \textbf{Type}                    & \textbf{Language}\\
1811 & Sense and Sensibility                   & Jane Austen                           & novel                   & English \\
1814 & Mansfield Park                          & Jane Austen                           & novel                   & English \\
1818 & Frankenstein                            & Mary Shelley                          & novel                   & English \\
1837 & The Pickwick Papers                     & Charles Dickens                       & novel                   & English \\
1839 & Nicholas Nickleby                       & Charles Dickens                       & novel                   & English \\
1839 & Oliver Twist                            & Charles Dickens                       & novel                   & English \\
1843 & A Christmas Carol                       & Charles Dickens                       & novella                 & English \\
1844 & Martin Chuzzlewit                       & Charles Dickens                       & novel                   & English \\
1847 & Jane Eyre                               & Charlotte Brontë                      & novel                   & English \\
1847 & Wuthering Heights                       & Emily Brontë                          & novel                   & English \\
1850 & David Copperfield                       & Charles Dickens                       & novel                   & English \\
1850 & The Scarlet Letter                      & Nathaniel Hawthorn                    & novel                   & English \\
1851 & Moby Dick                               & Herman Melville                       & novel                   & English \\
1852 & Uncle Tom's Cabin                       & Harriet Beecher Stowe                 & novel                   & English \\
1853 & Bleak House                   & Charles Dickens     & novel   & English \\
1856 & Madame Bovary                 & Gustave Flaubert (Eleanor Marx-Avelin)    & novel   & French  \\
1857 & Little Dorrit                           & Charles Dickens                       & novel                   & English \\
1859 & Adam Bede                     & George Eliot        & novel   & English \\
1861 & Great Expectations                      & Charles Dickens                       & novel                   & English \\
1865 & Alice's Adventures in Wonderland        & Lewis Carroll                         & novel                   & English \\
1866 & Crime and Punishment          & Fyodor Dostoevsky (Constance Garnett)   & novel   & Russian \\
1867 & War and Peace                           & Leo Tolstoy (Garnett)       & novel                   & Russian \\
1871 & Middlemarch                             & George Eliot                          & novel                   & English \\
1878 & Daisy Miller                            & Henry James                           & novella                 & English \\
1880 & Brothers Karamazov                      & Fyodor Dostoevsky (Garnett) & novel                   & Russian \\
1884 & Adventures of Huckleberry Finn          & Mark Twain                            & novel                   & English \\
1890 & The Picture of Dorian Gray              & Oscar Wilde                           & novel                   & English \\
1893 & Maggie: A Girl of the Streets & Stephen Crane       & novella & English \\
1895 & The Red Badge of Courage                & Stephen Crane                         & novel & English \\
1892 & Iola Leroy            & Frances Harper & novel & English \\
1897 & What Maisie Knew                        & Henry James                           & novel                   & English \\
1898 & The Turn of the Screw                   & Henry James                           & novella                 & English \\

1899 & The Awakening         & Kate Chopin    & novel & English \\
1900 & Sister Carrie                           & Theodore Dreiser                      & novel                   & English \\
1902 & The Sport of the Gods                   & Paul Laurence Dunbar                  & novel                   & English \\
1903 & The Ambassadors                         & Henry James                           & novel                   & English \\
1903 & The Call of the Wild                    & Jack London                           & novel                   & English \\
1903 & The Souls of Black Folk                 & W. E. B. Du Bois                      & collection (nonfiction) & English \\
1905 & House of Mirth                          & Edith Wharton                         & novel                   & English \\
1913 & O Pioneers!                             & Willa Cather                          & novel                   & English \\
1916 & A Portrait of the Artist as a Young Man & James Joyce                           & novel                   & English \\
1915 & The Rainbow           & D. H. Lawrence & novel & English \\
1918 & My Antonia                              & Willa Cather                          & novel                   & English \\
1920 & The Age of Innocence          & Edith Wharton       & novel   & English \\
1920 & This Side of Paradise         & F. Scott Fitzgerald & novel   & English \\
1922 & Jacob's Room                  & Virginia Woolf      & novel   & English \\
1922 & Swann's Way                   & Marcel Proust (C. K. Scott Moncrieff)       & novel   & French  \\
1925 & An American Tragedy                     & Theodore Dreiser                      & novel                   & English \\
1925 & Mrs Dalloway                            & Virginia Woolf                        & novel                   & English \\
1927 & To the Lighthouse                       & Virginia Woolf                        & novel                   & English \\
1928 & Lady Chatterly's Lover                  & D. H. Lawrence                        & novel                   & English \\
1932 & Brave New World                         & Aldous Huxley                         & novel                   & English \\
1936 & Gone with the Wind                      & Margaret Mitchell                     & novel                   & English \\
1931 & The Waves             & Virginia Woolf & novel & English\\
1945 & Animal Farm                   & George Orwell       & novel   & English\\
1949 & 1984                  & George Orwell  & novel & English \\
\bottomrule
\end{tabular}
  \caption{Primary sources from which training set windows were derived.}
  \label{appendix:training}
\end{table*}

\begin{table*}[h]
\footnotesize
\centering
\begin{tabular}{lllll}
\toprule
  & & \textbf{Validation Set} & &\\
 \midrule
\textbf{Year} & \textbf{Title}  & \textbf{Author (Translator)}   & \textbf{Type}             & \textbf{Language}\\
1815 & Emma       & Jane Austen   & novel   & English \\
1817 & Northanger Abbey   & Jane Austen    & novel   & English \\
1830 & The Red and the Black  & Stendhal (Horace B. Samuel)   & novel     & French  \\
1841 & Barnaby Rudge   & Charles Dickens   & novel   & English \\
1847 & Agnes Grey     & Anne Brontë         & novel        & English \\
1848 & The Tenant of Wildfell Hall    & Anne Brontë    & novel    & English \\
1854 & Hard Times       & Charles Dickens      & novel      & English \\
1859 & A Tale of Two Cities    & Charles Dickens  & novel    & English \\
1869 & Little Women    & Louisa May Alcott    & novel     & English \\
1877 & Anna Karenina   & Leo Tolstoy (Garnett)       & novel     & Russian \\
1883 & Treasure Island  & Robert Louis Stevenson & novel   & English \\
1898 & The War of the Worlds & H. G. Wells    & novel & English \\
1911 & Ethan Frome    & Edith Wharton     & novel   & English \\
1915 & The Song of the Lark   & Willa Cather      & novel      & English \\
1920 & Main Street     & Sinclair Lewis      & novel   & English \\
1922 & Babbitt         & Sinclair Lewis        & novel       & English \\
1922 & The Garden Party and Other Stories    & Katherine Mansfield     & collection (fiction)  & English \\
1925 & Arrowsmith      & Sinclair Lewis       & novel       & English \\
\midrule
 & & \textbf{Test Set} & & \\
\midrule
\textbf{Year} & \textbf{Title}                                   & \textbf{Author (Translator)}                 & \textbf{Type}                    & \textbf{Language}\\
1813 & Pride and Prejudice           & Jane Austen         & novel   & English \\
1817 & Persuasion                              & Jane Austen                           & novel                   & English \\
1899 & Heart of Darkness                       & Joseph Conrad                         & novella                 & English \\
1925 & The Great Gatsby                        & F. Scott Fitzgerald                   & novel                   & English \\
1934 & Tender Is the Night                     & F. Scott Fitzgerald                   & novel                   & English \\
\bottomrule
\end{tabular}
  \caption{Primary sources from which validation and test set windows were derived.}
  \label{appendix:val_test}
\end{table*}

\begin{table*}[t]
\begin{center}
\begin{tabular}{ rrrrrrrrrr } 
 \toprule
 \# of sents & \# instances & \multicolumn{6}{c}{recall@k} & mean rank & avg. \# candidates \\
 in quote &  & 1 & 3 & 5 & 10 & 50 & 100 \\
\midrule
1 & 3279 & 8.8 & 16.2 & 21.0 & 29.0 & 46.2 & 55.8 & 454.7 & 4913.0 \\
2 & 2028 & 11.0 & 21.5 & 27.4 & 35.6 & 55.5 & 65.2 & 337.6 & 4991.0 \\
3 & 1189 & 9.3 & 20.1 & 26.8 & 35.5 & 55.9 & 64.4 & 298.2 & 4873.7 \\
4 & 796 & 9.0 & 17.8 & 24.0 & 33.0 & 53.9 & 64.1 & 312.9 & 4753.5 \\
5 & 493 & 6.9 & 15.8 & 22.3 & 33.7 & 52.9 & 62.7 & 377.3 & 4549.9 \\
\bottomrule
\end{tabular}
\end{center}
\caption{A breakdown of performance by quote length in sentences of the performance of our best model, the dense retriever with 4 context sentences on each side. All numbers are on the test set of \name.}
\label{tab:best_model}
\end{table*}

\end{document}